\pdfoutput=1

\documentclass[11pt]{article}

\usepackage{EMNLP2022}

\usepackage{times}
\usepackage{latexsym}
\usepackage{enumitem}

\usepackage[T1]{fontenc}

\usepackage[utf8]{inputenc}

\usepackage{microtype}

\usepackage{inconsolata}

\usepackage{makecell}
\usepackage{multirow}
\usepackage{multicol}
\usepackage{tikz}
\usetikzlibrary{arrows}
\usepackage{float}
\usepackage{amsmath}
\usepackage{subcaption}

\title{Opening up Minds with Argumentative Dialogues}

\author{Youmna Farag$^1$ \ \ \ \ Charlotte O. Brand$^2$  \ \ \ \ Jacopo Amidei$^3$ \ \ \ \ Paul Piwek$^3$ \\ 
\textbf{Tom Stafford$^2$ \ \ \ \ Svetlana Stoyanchev$^4$ \ \ \ \ Andreas Vlachos$^1$} \\ $^1$University of Cambridge \ \ \ $^2$University of Sheffield  \\ $^3$The Open University \ \ \ $^4$Toshiba Cambridge Research Laboratory\\
\small{
 \texttt{\{yf273,av308\}@cam.ac.uk},} \
 \small{\texttt{\{c.brand,t.stafford\}@sheffield.ac.uk},} \\ \small{\texttt{\{paul.piwek,jacopo.amidei1\}@open.ac.uk},} \
\small{\texttt{svetlana.stoyanchev@crl.toshiba.co.uk}}}

\begin{document}
\maketitle
\begin{abstract}
 Recent research on argumentative dialogues has focused on
 persuading people to take some action, changing their stance on the topic of discussion, or winning debates. 
 In this work, we focus on argumentative dialogues that aim to open up (rather than change) people's minds to help them become more understanding to views that are unfamiliar or in opposition to their own convictions.
 To this end, we present a dataset of $183$ argumentative dialogues about $3$ controversial topics: veganism, Brexit and COVID-19 vaccination.
 The dialogues were collected using the Wizard of Oz approach, where wizards leverage a knowledge-base of arguments to converse with participants. 
 Open-mindedness is measured before and after engaging in the dialogue using a questionnaire from the psychology literature, and success of the dialogue is measured as the change in the participant’s stance towards those who hold opinions different to theirs.
We evaluate two dialogue models: a Wikipedia-based and an argument-based model. We show that while both models perform closely in terms of opening up minds, the argument-based model is significantly better on other dialogue properties such as engagement and clarity.

\end{abstract}

\section{Introduction}
Developing dialogue agents that are able to argue about different topics has been the focus of a lot of recent research. 
Typically, these agents engage in conversations with people with the aim of changing their opinions on a topic or winning debates. 
Accordingly, success of argumentative dialogue agents has been measured by their ability to convince people to take an action such as donating to a charity~\cite{wang-etal-2019-persuasion,Shi2020}, change their position on the subject of discussion~\cite{tan2016winning,prakken2020persuasive}, or attract more votes by the audience listening to their debates~\cite{zhang-etal-2016-conversational,slonim2021autonomous}. 
Other work has studied argumentation with the aim of reaching agreement~\cite{vecchi-etal-2021-towards,de-kock-vlachos-2021-beg}. 
Nonetheless, 
none of the previous works has studied dialogues in terms of their ability to stimulate open-minded thinking and help participants learn about views that are unfamiliar or in opposition to their own and become more tolerant towards people who hold these views. 

\begin{figure}[t]\scalebox{0.6}{
    \centering
    \includegraphics[clip, trim=0.1cm 20.5cm 5cm 0cm]{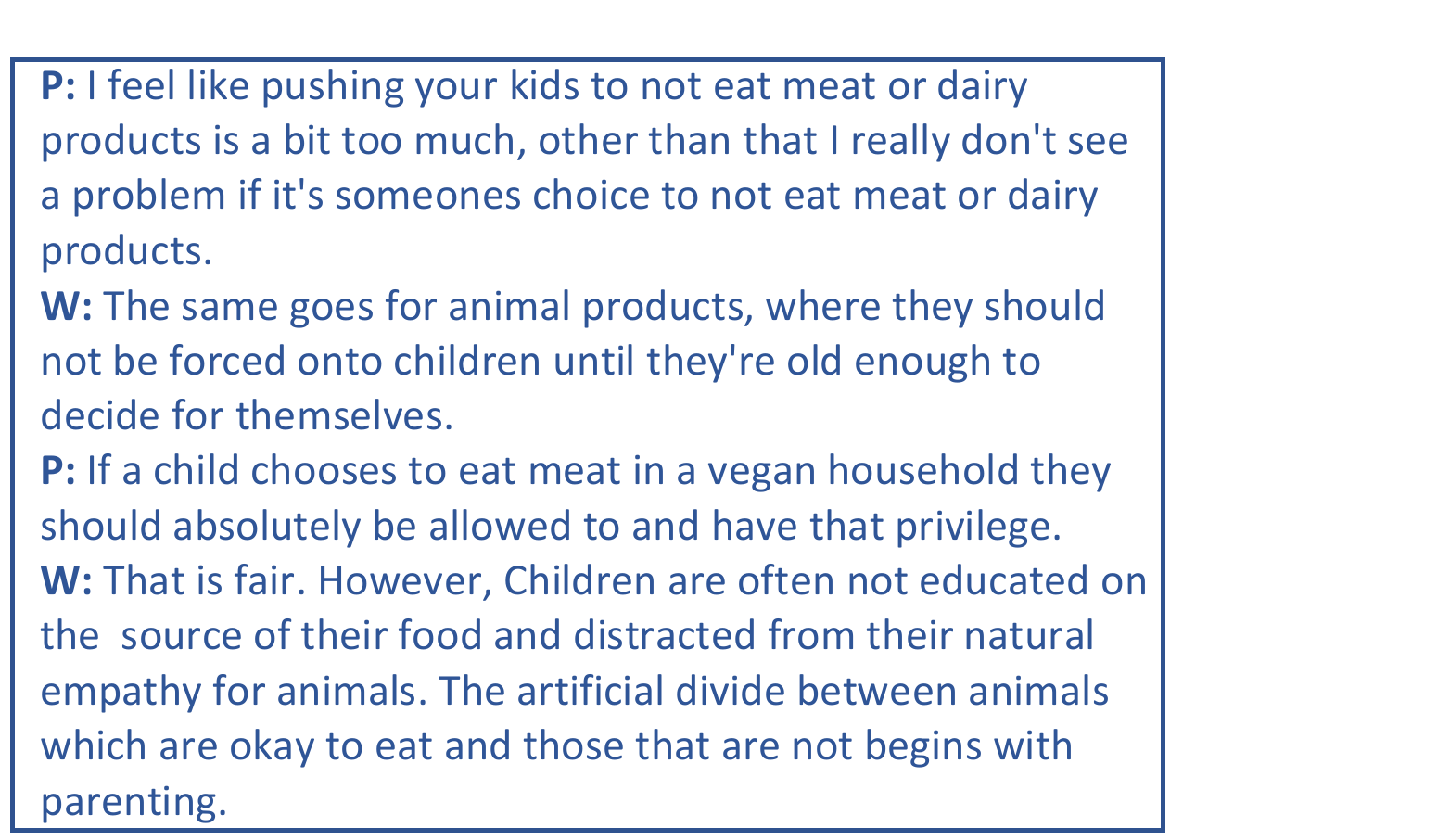}}
    \caption{A dialogue excerpt from our dataset about veganism between a participant (P) and a Wizard (W).}
    \label{fig:oum_example}
\end{figure}
Open-minded thinking has been motivated by many psychological studies.~\citet{haran2013role} showed that it correlates with information acquisition.

\citet{carpenter2018impact} demonstrated its importance for responsible behaviour on social media platforms.
More recently,~\citet{STANLEY2020104030} suggested that individual's negative views about their ideological opponents could partly be attributed to their lack of exposure to good arguments for these views. 
Motivated by research on open-minded thinking,
we propose to use argumentative dialogues to expose participants to different opinions about polarising topics with the aim of opening up their minds and increasing their tolerance towards views opposing their own. 
We collected $183$ dialogues about three controversial topics (veganism, Brexit and COVID-19 vaccination), using the Wizard of Oz (WoZ) approach~\citep{fraser1991simulating,bernsen2012designing}. The wizards utilised arguments sourced from publicly available debate platforms to chat with participants. Figure~\ref{fig:oum_example} shows an example from the dataset.

In order to evaluate open-mindedness, we follow the approach of~\citet{STANLEY2020104030}, and ask dialogue participants whether they believe people who hold views opposite to theirs have good reasons for their convictions.~\citet{STANLEY2020104030} argued that people who believe their ideological opponents have good reasons for their position are more likely to believe these opponents have good morals and intellectual capabilities. 
Therefore, we also ask participants about the intellectual capabilities and morality of people who hold views opposite to theirs. We refer to these questions as the \textit{opening up minds (OUM)} questions and detail them in Table~\ref{tab:stanley_questions}. We ask these questions before and after the dialogue and measure the change in the answers.  
Additionally, 
we ask participants to rate their experience (e.g., in terms of engagement, persuasiveness, frustration, etc.) and find no strong correlation between that and whether they have become more open-minded. These findings further highlight the distinction between dialogues aiming at opening up minds versus persuasiveness or engagement. 
To our knowledge, our dataset is the first dialogue corpus that aims at fostering open-minded thinking.\footnote{Dataset available at: \url{https://github.com/Youmna-H/OUMDials}}
Finally, we evaluate two dialogue models: a Wikipedia-based and an argument-based model, where the latter is fine-tuned on our dataset. Our results show that while both models perform closely in terms of opening up minds, the argument-based one is significantly 
better in other chat experience measures such as engagement and clarity.
\begin{table}[h!]
    \centering
    \scalebox{0.67}{
    \begin{tabular}{|c|l|}
    \hline
    Category & \multicolumn{1}{c|}{Questions} \\ \hline
    \multirow{2}{*}{\shortstack{Good \\Reasons}} & People who (have stance X)\\
    & have good reasons for (having stance x) \\ \hline
    \multirow{3}{*}{\shortstack{Intellectual \\ capabilities}} & People who (have stance X) are - Unintelligent \\
    & People who (have stance X) are - Irrational \\
    & People who (have stance X) are - Ignorant \\ \hline\multirow{3}{*}{Morality}
    & People who (have stance X) are - Unethical \\
    & People who (have stance X) are - Immoral \\
     & People who (have stance X) are - Of bad moral character \\ \hline
    \end{tabular}}
    \caption{OUM questions asked to participants before and after the conversation, following~\citet{STANLEY2020104030}. The part in brackets is substituted by the stance opposite to the participant's stance on the topic (e.g., people who are vegan/not vegan, people who voted Leave/Remain and people who have/have not had the COVID-19 vaccine).}
    \label{tab:stanley_questions}
\end{table}

\section{Related Work}
Several studies on argumentative dialogues have focused on persuasion.~\citet{tan2016winning} analysed the interactions on ChangeMyView (CMV) forums in order to understand the features that lead to persuasion. They described the original posters on CMV as ``open-minded'' if they changed their original view. In contrast, in our study an ``open-minded'' participant becomes more accepting to the opposite view, without necessarily changing theirs.~\citet{wang-etal-2019-persuasion} curated a dataset of dialogues where one participant tries to convince the other to make a donation. They studied different persuasion strategies that lead to dialogue success, which is measured by whether the participant actually made a donation. Following their work,~\citet{Shi2020} investigated the effect of chatbot identities on convincing people to make donations. Other work has focused on argumentative dialogues for debating such as Oxford-style Debates~\cite{zhang-etal-2016-conversational} and IBM's Project Debater~\cite{,slonim2021autonomous}. The goal of the participants  (humans or dialogue agents) in these debates is to win by convincing an audience with their arguments.

Recently, knowledge-based dialogue agents have attracted much attention in order to have more engaging dialogues and avoid knowledge hallucination, a typical issue in end-to-end chat models.
Numerous knowledge-bases have been utilised such as IMDB movie reviews~\cite{moghe-etal-2018-towards} or Wikipedia~\cite{zhou-etal-2018-dataset,dinan2019wizard}. For instance,~\citet{dinan2019wizard} used the WoZ approach to collect dialogues where wizards use sentences from Wikipedia to write their responses. These Wikipedia-based datasets have later been utilised to build knowledgeable dialogue agents~\cite{li-etal-2019-incremental,lian2019-ijcai,zhao-etal-2020-knowledge-grounded,Zhao2020Low-Resource,shuster2021retrieval}. Nonetheless, using arguments as a knowledge-base for dialogue agents has received less attention, with exception of, for example, ~\citet{prakken2020persuasive}, who developed a chatbot to persuade participants to accept that university fees should remain the same by selecting arguments from an argument graph using cosine similarity.

\begin{table}[t]
\centering
\scalebox{0.75}{
\begin{tabular}{|c|c|c|c|} 
\hline
 Topic & Veganism & Brexit  & Vaccination         \\ \hline
\#Dialogues & $73$ & $49$  & $61$ \\ \hline
Avg.\# Turns & $15.8\pm6.6$ & $14.5\pm5.6$ & $14.7\pm5.0$ \\ \hline
Argument base size & $5,384$ & $2,041$ & $1,982$ \\ 
\hline
\end{tabular}}
\caption{Statistics of the opening up minds (OUM) dataset.}  
\label{tab:stats}
\end{table}
\section{Wizard of Oz Data Collection}
\label{sec:woz}
We collect $183$ dialogues, using the WoZ approach, where a person (a wizard) plays the role of an agent and discusses a given topic with another person (a participant). 
Statistics of the collected dialogues are shown in Table~\ref{tab:stats}. In the remainder of this section, we discuss the dialogue collection process.

\paragraph{The wizards} 
We recruited $5$ postgraduate students from one of the author's university student job shop (a pool of students looking for research assistant work) to act as wizards.  
Each wizard is instructed to discuss a given topic for $15$-$20$ minutes with a participant to help them understand the other perspective on the topic being discussed rather than change their minds. 
More concretely, wizards are asked to use the most appropriate argument that best fits the conversation.
To assist them, an \textit{argument base} about the topic of discussion (see later in this section) is made available to them. Each argument is annotated with a \textit{pro} or \textit{con} stance relative to the topic.
After a participant's turn, TF-IDF scores are calculated between the participant's last utterance and each argument in the argument base,\footnote{If the last participant's utterance is less than $5$ words, we also consider the utterance before that.} and the  $50$ arguments with the highest scores are presented to the wizard to help them respond.
Wizards are encouraged to edit the arguments they select to make them flow more naturally with the conversation, or write their own responses from scratch if they can't find a good argument to use or want to ask questions. In order to further facilitate their task, 
wizards are also given a list of hedges and acknowledgments to use in their responses to make the conversation more natural and polite (e.g., ``I see what you mean, but...'', ``It could be argued...'', etc.), which previous research has found to be conducive to better conversations~\cite{YEOMANS2020131,de-kock-vlachos-2021-beg}.
The WoZ interface also allows the wizards to use keywords to search the whole argument base of the topic,
and to filter arguments by stance (pro/con). 

\paragraph{The participants} We recruited participants from Prolific\footnote{\url{https://prolific.co/}}. All  participants are fluent in English and have a Prolific acceptance rate of over $95\%$. Participants are asked to discuss the topic freely with the wizards, writing arguments and posing questions as they wish. Before the conversation, participants indicate their stance on the topic of discussion by answering whether: they are vegans (if the topic is veganism), they took at least one shot of the vaccine (if the topic is vaccination), or they voted leave or remain (if the topic is Brexit). 
According to their stance, they are asked about the people who hold the opposite stance; 
in particular, they indicate how much they disagree/agree with the OUM questions in Table~\ref{tab:stanley_questions} on a $7$-point Likert scale. They give their ratings before and after the dialogue.
Furthermore, participants are asked after the conversation about their chat experience by rating their chat on a $7$-point Likert scale to indicate how much it was: enjoyable, engaging, natural, clear, persuasive, confusing, frustrating, too complicated and boring (each measure is rated separately). They are also given the option to provide any other feedback about the conversation.  We include the instructions given to participants in Appendix~\ref{sec:appendix-hp}.

\paragraph{Argument base}
\label{argbase}
The arguments presented to the wizards are extracted from the online platform \textit{Kialo}\footnote{\url{https://www.kialo.com/}}. 
Arguments in Kialo are organised as a tree where the top node represents the main claim (the topic in our case). Each argument node in the tree is annotated with a \textit{pro} or \textit{con} stance based on whether it is for or against 
its parent argument node. In our WoZ platform, the arguments are labelled with their stances (pro or con) relative to the topic. As the nodes in Kialo are annotated with stances relative to their parent claim rather than the main claim/topic, we use a heuristic approach to calculate the stances relative to the topic. Specifically, we trace the argument tree from the topic node down to each child argument node and modify the stance of each child with the following assumptions:\footnote{We randomly select and manually inspect $60$ arguments from the three topics. We find that $43$ arguments were correctly classified by our approach, $13$ were neutral (i.e., neither pro nor con) and only $4$ were misclassified.}
\begin{itemize}[noitemsep,nolistsep]
    \item If an argument is pro the main topic, all its pro children will be pro the topic and all its con children will be con the topic.
    \item If an argument is con the main topic, all its pro children will be con the topic and all its con children will be pro the topic.
\end{itemize}
As vaccination had the lowest representation of arguments in Kialo, we augment the vaccination argument-base with  $479$ additional arguments written by participants who took part in a study examining anti-vaccination attitudes~\cite{brand2022using} and $108$ arguments sourced from a study examining the use of chatbots for changing vaccine attitudes~\cite{altay2021information,brand2022using}.

\begin{table}[t]
\centering
\scalebox{0.84}{
\begin{tabular}{|l|c|} 
\hline
\multicolumn{1}{|c|}{Wizard Action} & \% Percentage \\ \hline
Edit selected arg & $74.86$ \\
Use search terms & $68.77$ \\
Use stance filter & $71.76$ \\
Select arg from the top $10$ suggestions & $21.15$ \\
Use pro args & $47.40$ \\
Use con args & $52.60$ \\ \hline
\end{tabular}}
\caption{Percentage of the different (non-mutually exclusive) argument selection actions by  the wizards.} 
\label{tab:wizard-actions}
\end{table}

\paragraph{Wizard actions}
\label{sec:wizard_actions}
We find that wizards use arguments from the argument-base in $\approx 66\%$ of their responses.
In Table~\ref{tab:wizard-actions}, we detail statistics of different actions taken by the wizards when they select an argument from the argument base.
The table reveals that the wizards prefer to edit these arguments to fit the dialogue better ($74.86\%$ of the arguments were edited). Furthermore, they often use the search bar and the stance filter, instead of just selecting from the top arguments suggested by TF-IDF; they select an argument from the top $10$ suggestions only $21.15\%$ of the times. Finally, we notice that the wizards' use of pro and con arguments is balanced.

\begin{table*}[t]
\centering
\scalebox{0.65}{
\begin{tabular}{l|cccc|cccc|cccc} 
\hline
  \multirow{2}{*}{}& \multicolumn{4}{c|}{Good Reasons} & \multicolumn{4}{c|}{Morality} & \multicolumn{4}{c}{Intellectual Capabilities}       \\
& \%zero & \%$+$oum & \%$-$oum & overall   & \%zero & \%$+$oum & \%$-$oum & overall & \%zero & \%$+$oum & \%$-$oum & overall                                                \\ \hline
control-bot & $79.6$ & $12.2$ ($1.33$) & $8.2$ ($-1.0$) & $0.081$ & $69.4$ & $20.4$ ($0.97$) & $10.2$ ($-0.73$) & $0.122$ & $63.3$ & $26.5$ ($0.92$) & $10.2$ ($-0.93$) & $0.149$ \\

Wizards & $52.5$ & $35.8$ ($1.41$) & $11.7$ ($-1.32$) & $\textbf{0.35}^*$ & $56.5$  & $25.2$ ($1.05$)  & $18.3$ ($-0.73$) & $\textbf{0.131}$ & $53.8$ & $29.1$ ($1.1$) & $17.1$ ($-0.72$) & $\textbf{0.195}$ \\

Argu-bot & $65.3$ & $24$ ($1.5$) & $10.7$ ($-1.31$) & $0.22$  & $64$ & $18.7$ ($0.92$)  & $17.3$ ($-0.99$) & $0$  & $55.3$ & $27.3$ ($1.02$)  & $17.3$ ($-1.0$) &  $0.101$   \\

Wiki-bot & $61.3$ & $28$ ($1.4$)  & $10.7$ ($-1.56$) & $0.226$  & $66$ & $16$ ($0.79$)  & $18$ ($-0.9$) & $0$  & $60$ & $28$ ($0.76$)  & $12$ ($-0.5$) & $0.153$  \\
\hline
\end{tabular}}
\caption{The percentage of dialogues that have zero, positive or negative OUM scores in
the three OUM categories. `Overall' refers to the average of the dialogue's OUM scores for the respective category.
The numbers between brackets indicate the average OUM score. * indicates significance over control-bot using Welch t-test with $p<0.05$.} 
\label{tab:op_change}
\end{table*}

\section{Dialogue Models}
\label{sec:chatbots}
In this section, we describe the dialogue models 
for the task of opening up minds.
\paragraph{Wiki-bot}
We evaluate the Retrieval-Augmented Generation (RAG)-Sequence model~\cite{NEURIPS2020_6b493230} pre-trained on the Wizard-of-Wikipedia dataset~\cite{dinan2019wizard}. RAG-Sequence uses Wikipedia as a knowledge-base where a Dense Passage Retriever~\citep[DPR]{karpukhin-etal-2020-dense} is utilised to retrieve Wikipedia passages that are relevant to the dialogue history, then it uses BART~\cite{lewis-etal-2020-bart} to generate a dialogue response conditioned on the retrieved passages and the dialogue history.
We use the pre-trained model by~\citet{shuster-etal-2021-retrieval-augmentation}.\footnote{\url{https://parl.ai/projects/hallucination/}} Their approach uses beam search for decoding, however, we noticed that it suffers from repetition and therefore used nucleus sampling to remedy this. 

\paragraph{Argu-Bot}
We fine-tune the previously described wiki-bot on the OUM dataset (Section~\ref{sec:woz}). We split the dataset into $123$ dialogues for training, $15$ for validation and $45$ for testing.
Training is stopped when the validation perplexity doesn't improve for $5$ epochs. 
In order to accommodate for the nature of the dataset, we applied some adaptations to retrieval, training and generation as follows.
For retrieval:
\begin{itemize}[noitemsep,nolistsep]
 \item  Following the wizards' experiments, we use Kialo arguments, instead of Wikipedia, as the knowledge-base for the retrieval model.
 \item  We use BM25 instead of DPR for retrieval as initial experiments showed that DPR is more suited for Wikipedia but not suitable for argument retrieval.\footnote{Lower-casing, stemming and removing stop words are applied to the arguments and dialogue history before retrieval.}
\item We assume that the arguments used by the wizards in the training data are good arguments and accordingly increase their scores by $1$ if they are retrieved by BM25.
 \item   We make use of the search terms the wizards used to find arguments (Section~\ref{sec:woz}) and compile a list of ``important terms''. We increase the scores of retrieved arguments by $1$ if they include any of these terms.
 \item  We pay more attention to the recent dialogue history by increasing the scores of the retrieved arguments by $1$ if they have overlapping terms with the  participant's last utterance.
\end{itemize}
For training:
\begin{itemize}[noitemsep,nolistsep]
\item At any point in the dialogue, the model is optimised to generate a response similar to the wizard's. If the wizard used an argument to write their response, the model uses this ``gold'' wizard argument instead of retrieving one by BM25. If, however, the wizard did not use an argument, the model uses the top one retrieved by BM25. By doing this, the model can learn how arguments are edited to compose responses the way wizards do. During testing we only use the top BM25 argument.
\item We compute a loss function for the model to learn how much to use arguments in generation, similar to work in abstractive summarisation~\cite{see-etal-2017-get}. At any turn $t$ in the dialogue, the model learns a generation probability ($pgen_t \in [0,1]$) conditioned on the participant's last utterance $h_t$:
\begin{equation}
    pgen_t = \sigma(W\cdot h_t + b)
\end{equation}
where $pgen_t$ is optimized to be $0$ if the wizard used an argument to generate the response and $1$ otherwise. During inference, the probability of generating a response sequence $y$ is calculated by:
\begin{align}
\label{eqn:loss}
   p(y|x) = \prod_i (pgen_t \cdot p(y_i|x,y_{i-1}) \\\nonumber + (1-pgen_t) \cdot p(y_i|x,z,y_{i-1}))
\end{align}
where $x$ is the dialogue history and $z$ is the retrieved argument. 
\end{itemize}
Finally, for generation, we re-rank the candidate responses generated by nucleus sampling w.r.t.\ their similarity to the retrieved argument and dissimilarity to the previously generated utterances (to avoid repetition). In order to achieve this, we compute the BLEU score between each candidate response and the retrieved argument and the negative of the BLEU score between each candidate and the previously generated utterances, then re-rank the candidates using the average of these two scores. 

\paragraph{Control-bot}
We use a control condition in our experiment to verify whether participants change their ratings for the OUM questions due to discussing the topic, or other reasons such as demand effect (i.e., they think they are required to change their ratings positively). To this end, we evaluate a `chitchat' chatbot and instruct the participants to chat about their holidays/weekends. We use the same format of before and after questions as in the wizards study about the $3$ topics (veganism, Brexit and vaccination). For instance, in an experiment about veganism, a vegan participant is first asked about their views about non-vegans, then they talk with the chatbot about their holidays, then after the chat they are asked again about their views about non-vegans. We use a Polyencoder model trained on the ConvAI2 dialogues~\cite{Humeau2020Poly-encoders:} and we refer to this chatbot as \textit{control-bot}.

\section{Evaluation}
We evaluate the models described in Section~\ref{sec:chatbots} using the same setup as in Section~\ref{sec:woz} but by replacing the wizards with one of the models, and limiting the chat time to $10$-$15$ instead of $15$-$20$ minutes as the models are much faster than the wizards. We collect $150$ dialogues for each of the argu-bot and wiki-bot ($60$ for veganism, $45$ for Brexit and $45$ for vaccination) and $50$ dialogues for the control-bot ($20$ for veganism, $15$ for Brexit and $15$ for vaccination). In the remainder of this section, we present analysis of open-mindedness and chat experience for the wizards and the dialogue models.

\subsection{Opening-up Minds}
\label{sec:oum-scores}
As discussed in Section~\ref{sec:woz}, we ask the participants a set of OUM questions before and after the dialogue in order to evaluate the impact of the dialogue on changing their attitude towards those holding opinions different to theirs. If we  
ignore the dialogues where the participants did not respond to the questions after the dialogues, the number of dialogues with OUM question annotations becomes $120$ for the wizards, $150$ for argu-bot, $150$ for wiki-bot and $50$ for control-bot. For each dialogue, we calculate three \textit{OUM scores} corresponding to the three question categories defined in Table~\ref{tab:stanley_questions}. Each OUM score is calculated as the difference between the ratings before and after the dialogue. As the morality and intellectual capabilities categories contain three questions each, the score for the category is the average of the changes in its sub-questions.
We note that due to the different phrasing of the OUM questions, an increase in the rating for the “good reasons” question denotes a positive change, whereas a decrease in the ratings for “intellectual capabilities” and “morality” questions denotes a positive change.
We categorise the dialogues according to their OUM scores into $3$ classes: \textit{zero change}: where the score $= 0$, \textit{+oum change}: where the score $> 0$ and \textit{-oum change}: where the score $< 0$. We show in Table~\ref{tab:op_change}, the percentage of dialogues in each class and the average OUM score per class.
We also report the overall score as the average of the OUM scores of all the dialogues for each OUM question category. This overall score helps report the model's success with a single number.\footnote{For example, if $x$ is the number of collected dialogues and $x/2$ dialogues have OUM score = $1$ and the other $x/2$ have OUM score = $-1$, for the good reasons question, the overall score of the model for good reasons will be zero.}

\begin{figure}
\scalebox{0.36}{
    \centering
    \includegraphics{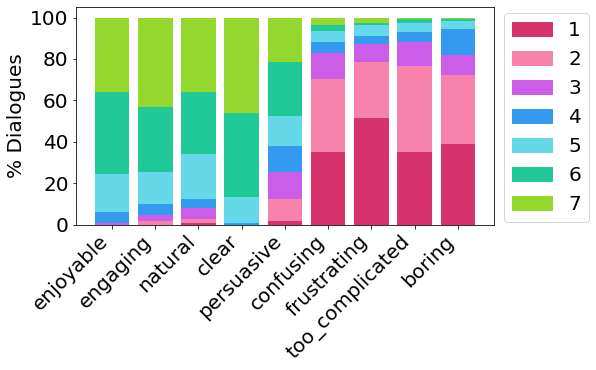}}
    \caption{The graph depicts the ratings of wizards' dialogues in terms of chat experience. The y-axis corresponds to the proportion of the dialogues, the x-axis corresponds to aspects of chat experiences, and the different colors refer to the ratings on the $7$-point Likert scale, where $1$=strongly disagree and $7$=strongly agree.}
    \label{fig:experience}
\end{figure}

\subsubsection{Wizards}
The results in Table~\ref{tab:op_change} demonstrate the success of the wizards' dialogues in opening up participants' minds, particularly with the good reasons category ($35.8\%$ of the dialogues resulted in a positive OUM change). We find that despite the fact that for each question category most participants have zero change, which is expected given the relative brevity of the dialogues, the number of participants who have a positive change in their attitude (+oum) is substantially larger than those who have negative change (-oum). Even when the percentage of dialogues with negative scores is relatively high (e.g., $18.3\%$ in the morality category), the average OUM score is smaller than in the positive dialogues (e.g., $-0.73$ vs $1.05$ with the morality category), and all the categories have a positive overall score.
Additionally, we find that the percentage of the dialogues with zero change in the control-bot is higher than the wizards in all question categories, which demonstrates the effect of conversing with wizards in comparison to the control condition.
Furthermore, the wizards are consistently better than all the models in all question categories in terms of overall score, with a statistically significant difference over the control-bot in the good reasons category. In general, we notice that participants tend to become more open-minded about the good reasons people might have for their stances (with overall score $ = 0.35$), which reflects the nature of the argumentative dialogues and the wizards success in finding good arguments that stimulate open-minded thinking. 

On the other hand, the difference between the wizards and the control-bot is less obvious with the morality and intellectual capabilities questions. We investigate this and take a closer look at the OUM ratings \textbf{before} the dialogue. We find that $39.4\%$ of the participants strongly agree/agree that their opponents have good reasons for their convictions, while $44.5\%$ and $56.7\%$ strongly disagree/disagree that their opponents have low intellectual capabilities or morality respectively. When we look at the most open-minded ratings, we find that only $14.7\%$ of the participants strongly agree that their opponents have good reasons for their position, while $24.1\%$ and $30.5\%$ strongly disagree that their opponents have low intellectual capabilities or morality respectively. 
This shows that regarding the intellectual capabilities and morality categories, particularly the latter, participants come (before the dialogue) with a more open mind than in the good reasons category, and while they might not completely agree with the reasons their opponents have, they are less harsh in their judgement of the morality of these opponents. Therefore, the dialogues have more room to improve the rating of the reasons for the opposite view.  
The results of the morality and intellectual capabilities also suggest that there is room for development of novel measures which provide further insight into the mechanisms behind changes in open-mindedness.

\begin{table}[t]
\centering
\scalebox{0.7}{
\begin{tabular}{|c|c||c|c||c|} 
\hline
    & Wizards & Argu-bot & Wiki-bot & control-bot  \\ \hline
enjoyable & $\textbf{6.05}$ & $5.13^*$ & $4.77$ & $4.24$ \\
engaging & $\textbf{6.02}$ & $5.09^{**}$ & $4.57$ & $4.02$ \\
natural & $\textbf{5.77}$ & $3.81$ & $3.46$ & $2.96$ \\
clear & $\textbf{6.32}$ & $5.276^{***}$ & $4.66$ & $3.92$ \\
persuasive & $\textbf{4.92}$ & $4.33^{***}$ & $3.71$ & $3.16$\\
consistent & - & $\textbf{4.56\textsuperscript{**}}$ & $4.0$ & $3.59$\\
knowledgeable & - & $\textbf{5.32\textsuperscript{***}}$ & $4.45$ & $3.06$\\ \hline
confusing & $\textbf{2.33}$ & $3.58^{***}$ & $4.75$ & $4.82$ \\
frustrating & $\textbf{1.98}$ & $3.09^{***}$ & $3.79$ & $4.27$ \\
too complicated & $\textbf{2.11}$ & $2.93$ & $3.03$ & $2.47$\\
boring & $\textbf{2.15}$ & $3.27$ & $3.52$ & $4.08$ \\ \hline
\end{tabular}}
\caption{Average ratings for chat experiences on a $7$-point Likert scale. In the top rows the higher the score is the better while in the bottom rows the lower the score is the better.
Statistical significance is calculated using the Welch t-test between argu-bot and wiki-bot where *** $p<0.001$, ** $p<0.01$ and *$p<0.05$.}
\label{tab:all_experience}
\end{table}

We further investigate the correlation of features of the wizard dialogues with the success of these dialogues in opening-up minds with respect to the good reasons question. For this purpose we calculate Spearman’s rank correlation coefficient ($\rho$) between the OUM scores for the good reasons question and the following dialogue features:
\begin{itemize}[noitemsep,nolistsep]
    \item Length-related features: dialogue length computed as the total number of turns in the dialogue, proportion of wizard turns, and proportion of participant turns.
    \item Proportion of questions asked by the wizard to the total number of sentences in their turns. We use Stanford CoreNLP parser~\cite{manning-etal-2014-stanford} for question identification.
    \item Proportion of utterances containing arguments selected from the argument base to the total number of wizard turns.
    \item Proportion of edited arguments w.r.t. all the arguments selected and used by the wizards.
    \item Ratio between pro and con arguments used by the wizard.
    \item Frequency of politeness markers~\cite{danescu-niculescu-mizil-etal-2013-computational} utilised by the wizard such as greetings, hedging and subjunctives. We use Convokit~\cite{chang-etal-2020-convokit} to identify politeness markers and normalise each marker by the number of sentences written by the wizard.
\end{itemize}
Our analysis reveals very weak to negligible correlations between the OUM scores for good reason and any of these features.\footnote{See Appendix~\ref{sec:features-app} for the full table of correlations.}
 The features with the strongest correlations are two of the politeness features: the use of positive lexicon ($\rho=0.18$) and the use of subjunctives\footnote{Example of a subjunctive: ``Would you agree that eating meat is not inherently bad''.} ($\rho=-0.18$). While using positive words fosters a positive attitude towards the participant (e.g., by acknowledging their ``good'' points), it is not clear why there is negative correlation between subjunctives and OUM scores.
 
\subsubsection{Models}
Table~\ref{tab:op_change} shows that both the wiki-bot and the argu-bot have a higher overall score than control-bot in terms of the good reasons question. This further demonstrates the ability of the two models to positively change people's attitudes regarding the reasons their opponents have. Nevertheless, this change diminishes in the morality and intellectual categories, which aligns with our findings from the comparison between the wizards and control-bot. 
Additionally, despite the fact that the efficacy of using arguments was demonstrated by the wizards' performance that surpassed all the chat models (with good reasons overall score $=0.35$), the model that leverages arguments (argu-bot) performs on par with the one that uses Wikipedia (wiki-bot), with good reasons overall score $\approx0.22$. 
We conjecture that the impact of argu-bot could be improved by: (1) training a retriever model on the OUM dataset to learn how to retrieve arguments similarly to the wizards (2) investigating different loss functions as we find that the model heavily relies on the arguments instead of also asking questions like the wizards do.  
In Table~\ref{tab:eval-examples}, we give an example of vaccination dialogues that had a positive OUM score for good reasons. The example shows that argu-bot is more able to engage in the conversation and talk about COVID-19 vaccines with more knowledge, whereas wiki-bot says it does not know about COVID-19 and deviates to other vaccines and diseases. Nonetheless, this wiki-bot dialogue had a positive change in the good reasons category which shows the potential of dialogue agents for opening-up minds.

\begin{table}[t]
\centering
\scalebox{0.8}{
\begin{tabular}{c|c} 
\hline
Experience & $\rho$ \\ \hline
enjoyable &  $0.01$ \\
engaging &  $-0.01$\\
natural &  $0.07$ \\
clear &   $0.04$\\
persuasive & $0.16$\\
confusing &  $0.02$\\
frustrating &  $-0.11$\\
too complicated &  $-0.19$\\
boring &  $-0.16$\\ \hline
\end{tabular}}
\caption{Spearman's correlation ($\rho$) between OUM scores for the good reason question and ratings for chat experience in the wizards' dialogues.} 
\label{tab:experience-oum}
\end{table}

\begin{table*}[h!]
\centering
\scalebox{0.77}{
\begin{tabular}{c}
\hline
Wiki-bot  \\ \hline
\multicolumn{1}{m{17cm}}{M: Hello, What do you think about COVID-19 vaccination?}\\
\multicolumn{1}{m{17cm}}{P: I think, they are good for us.} \\
\multicolumn{1}{m{17cm}}{M: I don't know about COVID-19, but I do know that the WHO recommends getting vaccinated against typhoid and shingles (common cold).} \\
\multicolumn{1}{m{17cm}}{P: yes, that's true} \\
\multicolumn{1}{m{17cm}}{M: THomid Typhoid vaccine is actually derived from the negative causative agents in the bacterium Streptococcus aureus.} \\
\multicolumn{1}{m{17cm}}{P: I have to say, that i didn't heard about it.} \\ \hline
Argu-bot  \\ \hline
\multicolumn{1}{m{17cm}}{M: Hello, What do you think about COVID-19 vaccination?}\\
\multicolumn{1}{m{17cm}}{P: I think it's important for people to get immunized against COVID-19} \\
\multicolumn{1}{m{17cm}}{M: What is your opinion of some Covid-19 vaccines will contain adjuvants. Adjuvants increase the immune response, and therefore reduce the amount of antigen that must be used in each dose of vaccine. The adjuvents therefore help to ensure that a large quantity of doses can be produced quickly.} \\
\multicolumn{1}{m{17cm}}{P: Sounds like they help with speeding up the vaccination process, so they must be good} \\
\multicolumn{1}{m{17cm}}{M: Those are good points, although some people feel the speed in which the vaccine was made has made them nervous about taking it. They don't believe that a vaccine made this quickly would not be safe and they think corners must have been cut  to make the vaccines work quickly.} \\
\multicolumn{1}{m{17cm}}{P: Well, if they do not take the vaccine, they risk their health and their close ones' health. Even if the vaccine is not to be fully trusted, what could those people lose that's worse than their life and their family's life?} \\
 \hline

\end{tabular}}
\caption{Excerpts from dialogues with the argu-bot and the wiki-bot about COVID-19 vaccination, where `M:' and `P:' mark the model and the participant turns respectively. Both dialogues achieved a positive change in the good reasons category.} 
\label{tab:eval-examples}
\end{table*}

\subsection{Chat Experience}
After the dialogue, participants are asked to rate their chat experience on a scale from $1$ to $7$ in terms of how much it was: enjoyable, engaging, natural, confusing, frustrating, clear, persuasive, too complicated and boring. With the dialogue models we add two more metrics: consistent and knowledgeable. We present the chat experience average ratings in Table~\ref{tab:all_experience}.

\subsubsection{Wizards}
Table~\ref{tab:all_experience} shows that the wizards surpass all the other models in terms of chat experience.
In Figure~\ref{fig:experience}, we plot the distribution of chat experience ratings. It is clear from the figure that participants mostly strongly agree/agree with the positive experiences (e.g., enjoyable) and mostly strongly disagree/disagree with the negative ones (e.g., frustrating), which is another sign of wizards' success. 

We further investigate the correlation between chat experience ratings and the OUM scores of wizard dialogues for the good reason question (Section~\ref{sec:oum-scores}). Based on the results in Table~\ref{tab:experience-oum}, we can see that there is no strong correlation between the scores and the different experiences; there is very weak negative correlation with some of the bad experiences (e.g., $\rho=-0.19$ for too complicated) and very weak positive correlation with some of the good experiences (e.g., $\rho=0.16$ for persuasive). These results show that participants can still enjoy the conversation and have a positive experience even if they did not change their position. The weak correlation between OUM scores and persuasiveness further demonstrates the difference between persuading someone and opening up their minds about the different opinions which motivates building dialogue systems that foster open-mindedness.
Participants are also given the option to write any other feedback about the conversation. We find that all the feedback to the wizards was positive and included sentences like: ``The bot is much more nice than the average human who asks these kind of questions'',\footnote{We note that the participants do not know they are talking to a human, as this is how WoZ experiments are conducted.} ``It has opened my eyes to the possibilities of vegan lifestyle and their benefits '' and ``This study was very enjoyable and fun. I learnt a lot from it.''.

\subsubsection{Models}
Table~\ref{tab:all_experience} reveals that argu-bot surpasses wiki-bot in all chat experience metrics and is significantly better on $8$ of them. The high performance on chat experience is important to build real-life dialogue models that aim to open up minds. This is because, while in our experiments participants were clearly asked to stay at least $10$ minutes in the chatroom and otherwise their experiment gets rejected, in real-life, this restriction does not apply and therefore participants need to find the chatbot enjoyable and engaging in order to continue chatting with it. We also find that the feedback argu-bot received is more positive than wiki-bot and included sentences like: ``Interesting and made me think about my choices :)'' or ``I really liked this one. Chatbots are a clever way of engaging into a topic.''. However, participants were more critical than with the wizards and added comments like: ``One time bot answered two times exactly the same answer. It should be improved, however the overall impression of it is fine :)'' and ``I think the responses of the chatbot didn't answered my questions, they were missing the point.''. Feedback about wiki-bot included: ``The chatbot was changing the topic and not relating to my sentences''.\footnote{We include more detailed figures for chat experience in Appendix~\ref{sec:bots-experience}.}

\section{Conclusion}
We presented a dataset of argumentative dialogues for opening-up minds and showed its success in positively changing participants attitudes regarding the reasons people have for their opposing views. However, this impact was lower with regard to the morality and intellectual capabilities measures, which warrants further study to these measures. We evaluated two dialogue models: a Wikipedia-based and an argument-based one and showed that while they both perform closely in terms of opening up minds, the argument-based model is more successful in providing a good chat experience.

\section*{Limitations}
\begin{itemize}[noitemsep,nolistsep]
    \item It would be useful to train a neural retriever model for argu-bot to learn to select arguments like the wizards (instead of using BM25), but this requires collecting more wizard dialogues.
    \item Collecting more dialogues with wizards is an expensive process as it requires training more wizards and paying both wizards and participants. 
    \item Our study involves measuring individuals on how open-minded they are with respect to a position they are opposed to. While we rely on recent research in psychology for this, we acknowledge that such measurements are difficult and more research is needed in this direction.
    \item We only studied the effects of the dialogues on the participants immediately after they were held, but did not check whether the effect was long-term or short-lived.
\end{itemize}

\section*{Ethics Statement}
We have obtained approval for dialogue collection from the ethics review board of the University of Sheffield. Wizards were paid £$14.86$ an hour, in line with university regulations for paying research assistants.
Participants were paid £$9$ an hour, above Prolific’s minimum of £$7$ an hour and in line with the UK national living wage for 2021. Participants were informed that the statements provided by the chatbot they will interact with have not been fact-checked. All participants' personal information is anonymized. 

\section*{Acknowledgements}
This research was carried out as part of the Opening Up Minds project, funded by the UK Engineering and Physical Sciences Research Council under the linked grants EP/T024666/1, EP/T023414/1 and EP/T023554/1. We would also like to thank Haoming Wang for his help with developing the interface used for the Wizard of Oz data collection.

\bibliography{anthology,custom}

\begin{thebibliography}{33}
\expandafter\ifx\csname natexlab\endcsname\relax\def\natexlab#1{#1}\fi

\bibitem[{Altay et~al.(2021)Altay, Hacquin, Chevallier, and
  Mercier}]{altay2021information}
Sacha Altay, Anne-Sophie Hacquin, Coralie Chevallier, and Hugo Mercier. 2021.
\newblock Information delivered by a chatbot has a positive impact on covid-19
  vaccines attitudes and intentions.
\newblock \emph{Journal of Experimental Psychology: Applied}.

\bibitem[{Bernsen et~al.(2012)Bernsen, Dybkj{\ae}r, and
  Dybkj{\ae}r}]{bernsen2012designing}
Niels~O Bernsen, Hans Dybkj{\ae}r, and Laila Dybkj{\ae}r. 2012.
\newblock \emph{Designing interactive speech systems: From first ideas to user
  testing}.
\newblock Springer Science \& Business Media.

\bibitem[{Brand and Stafford(2022)}]{brand2022using}
Charlotte~O Brand and Tom Stafford. 2022.
\newblock Using dialogues to increase positive attitudes towards covid-19
  vaccines in a vaccine-hesitant uk population.
\newblock \emph{Royal Society Open Science}.

\bibitem[{Carpenter et~al.(2018)Carpenter, Preotiuc-Pietro, Clark, Flekova,
  Smith, Kern, Buffone, Ungar, and Seligman}]{carpenter2018impact}
Jordan Carpenter, Daniel Preotiuc-Pietro, Jenna Clark, Lucie Flekova, Laura
  Smith, Margaret~L Kern, Anneke Buffone, Lyle Ungar, and Martin Seligman.
  2018.
\newblock The impact of actively open-minded thinking on social media
  communication.
\newblock \emph{Judgment \& Decision Making}, 13(6).

\bibitem[{Chang et~al.(2020)Chang, Chiam, Fu, Wang, Zhang, and
  Danescu-Niculescu-Mizil}]{chang-etal-2020-convokit}
Jonathan~P. Chang, Caleb Chiam, Liye Fu, Andrew Wang, Justine Zhang, and
  Cristian Danescu-Niculescu-Mizil. 2020.
\newblock \href {https://aclanthology.org/2020.sigdial-1.8} {{C}onvo{K}it: A
  toolkit for the analysis of conversations}.
\newblock In \emph{Proceedings of the 21th Annual Meeting of the Special
  Interest Group on Discourse and Dialogue}, pages 57--60, 1st virtual meeting.
  Association for Computational Linguistics.

\bibitem[{Danescu-Niculescu-Mizil et~al.(2013)Danescu-Niculescu-Mizil, Sudhof,
  Jurafsky, Leskovec, and
  Potts}]{danescu-niculescu-mizil-etal-2013-computational}
Cristian Danescu-Niculescu-Mizil, Moritz Sudhof, Dan Jurafsky, Jure Leskovec,
  and Christopher Potts. 2013.
\newblock \href {https://aclanthology.org/P13-1025} {A computational approach
  to politeness with application to social factors}.
\newblock In \emph{Proceedings of the 51st Annual Meeting of the Association
  for Computational Linguistics (Volume 1: Long Papers)}, pages 250--259,
  Sofia, Bulgaria. Association for Computational Linguistics.

\bibitem[{De~Kock and Vlachos(2021)}]{de-kock-vlachos-2021-beg}
Christine De~Kock and Andreas Vlachos. 2021.
\newblock \href {https://doi.org/10.18653/v1/2021.eacl-main.173} {{I} beg to
  differ: A study of constructive disagreement in online conversations}.
\newblock In \emph{Proceedings of the 16th Conference of the European Chapter
  of the Association for Computational Linguistics: Main Volume}, pages
  2017--2027, Online. Association for Computational Linguistics.

\bibitem[{Dinan et~al.(2019)Dinan, Roller, Shuster, Fan, Auli, and
  Weston}]{dinan2019wizard}
Emily Dinan, Stephen Roller, Kurt Shuster, Angela Fan, Michael Auli, and Jason
  Weston. 2019.
\newblock {W}izard of {W}ikipedia: Knowledge-powered conversational agents.
\newblock In \emph{Proceedings of the International Conference on Learning
  Representations (ICLR)}.

\bibitem[{Fraser and Gilbert(1991)}]{fraser1991simulating}
Norman~M Fraser and G~Nigel Gilbert. 1991.
\newblock Simulating speech systems.
\newblock \emph{Computer Speech \& Language}, 5(1):81--99.

\bibitem[{Haran et~al.(2013)Haran, Ritov, and Mellers}]{haran2013role}
Uriel Haran, Ilana Ritov, and Barbara~A Mellers. 2013.
\newblock The role of actively open-minded thinking in information acquisition,
  accuracy, and calibration.
\newblock \emph{Judgment and Decision Making}, 8:188--201.

\bibitem[{Humeau et~al.(2020)Humeau, Shuster, Lachaux, and
  Weston}]{Humeau2020Poly-encoders:}
Samuel Humeau, Kurt Shuster, Marie-Anne Lachaux, and Jason Weston. 2020.
\newblock \href {https://openreview.net/forum?id=SkxgnnNFvH} {Poly-encoders:
  Architectures and pre-training strategies for fast and accurate
  multi-sentence scoring}.
\newblock In \emph{International Conference on Learning Representations}.

\bibitem[{Karpukhin et~al.(2020)Karpukhin, Oguz, Min, Lewis, Wu, Edunov, Chen,
  and Yih}]{karpukhin-etal-2020-dense}
Vladimir Karpukhin, Barlas Oguz, Sewon Min, Patrick Lewis, Ledell Wu, Sergey
  Edunov, Danqi Chen, and Wen-tau Yih. 2020.
\newblock \href {https://doi.org/10.18653/v1/2020.emnlp-main.550} {Dense
  passage retrieval for open-domain question answering}.
\newblock In \emph{Proceedings of the 2020 Conference on Empirical Methods in
  Natural Language Processing (EMNLP)}, pages 6769--6781, Online. Association
  for Computational Linguistics.

\bibitem[{Lewis et~al.(2020{\natexlab{a}})Lewis, Liu, Goyal, Ghazvininejad,
  Mohamed, Levy, Stoyanov, and Zettlemoyer}]{lewis-etal-2020-bart}
Mike Lewis, Yinhan Liu, Naman Goyal, Marjan Ghazvininejad, Abdelrahman Mohamed,
  Omer Levy, Veselin Stoyanov, and Luke Zettlemoyer. 2020{\natexlab{a}}.
\newblock \href {https://doi.org/10.18653/v1/2020.acl-main.703} {{BART}:
  Denoising sequence-to-sequence pre-training for natural language generation,
  translation, and comprehension}.
\newblock In \emph{Proceedings of the 58th Annual Meeting of the Association
  for Computational Linguistics}, pages 7871--7880, Online. Association for
  Computational Linguistics.

\bibitem[{Lewis et~al.(2020{\natexlab{b}})Lewis, Perez, Piktus, Petroni,
  Karpukhin, Goyal, K\"{u}ttler, Lewis, Yih, Rockt\"{a}schel, Riedel, and
  Kiela}]{NEURIPS2020_6b493230}
Patrick Lewis, Ethan Perez, Aleksandra Piktus, Fabio Petroni, Vladimir
  Karpukhin, Naman Goyal, Heinrich K\"{u}ttler, Mike Lewis, Wen-tau Yih, Tim
  Rockt\"{a}schel, Sebastian Riedel, and Douwe Kiela. 2020{\natexlab{b}}.
\newblock \href
  {https://proceedings.neurips.cc/paper/2020/file/6b493230205f780e1bc26945df7481e5-Paper.pdf}
  {Retrieval-augmented generation for knowledge-intensive nlp tasks}.
\newblock In \emph{Advances in Neural Information Processing Systems},
  volume~33, pages 9459--9474. Curran Associates, Inc.

\bibitem[{Li et~al.(2019)Li, Niu, Meng, Feng, Li, and
  Zhou}]{li-etal-2019-incremental}
Zekang Li, Cheng Niu, Fandong Meng, Yang Feng, Qian Li, and Jie Zhou. 2019.
\newblock \href {https://doi.org/10.18653/v1/P19-1002} {Incremental transformer
  with deliberation decoder for document grounded conversations}.
\newblock In \emph{Proceedings of the 57th Annual Meeting of the Association
  for Computational Linguistics}, pages 12--21, Florence, Italy. Association
  for Computational Linguistics.

\bibitem[{Lian et~al.(2019)Lian, Xie, Wang, Peng, and Wu}]{lian2019-ijcai}
Rongzhong Lian, Min Xie, Fan Wang, Jinhua Peng, and Hua Wu. 2019.
\newblock \href {https://doi.org/10.24963/ijcai.2019/706} {Learning to select
  knowledge for response generation in dialog systems}.
\newblock In \emph{Proceedings of the Twenty-Eighth International Joint
  Conference on Artificial Intelligence, {IJCAI-19}}, pages 5081--5087.
  International Joint Conferences on Artificial Intelligence Organization.

\bibitem[{Manning et~al.(2014)Manning, Surdeanu, Bauer, Finkel, Bethard, and
  McClosky}]{manning-etal-2014-stanford}
Christopher Manning, Mihai Surdeanu, John Bauer, Jenny Finkel, Steven Bethard,
  and David McClosky. 2014.
\newblock \href {https://doi.org/10.3115/v1/P14-5010} {The {S}tanford
  {C}ore{NLP} natural language processing toolkit}.
\newblock In \emph{Proceedings of 52nd Annual Meeting of the Association for
  Computational Linguistics: System Demonstrations}, pages 55--60, Baltimore,
  Maryland. Association for Computational Linguistics.

\bibitem[{Moghe et~al.(2018)Moghe, Arora, Banerjee, and
  Khapra}]{moghe-etal-2018-towards}
Nikita Moghe, Siddhartha Arora, Suman Banerjee, and Mitesh~M. Khapra. 2018.
\newblock \href {https://doi.org/10.18653/v1/D18-1255} {Towards exploiting
  background knowledge for building conversation systems}.
\newblock In \emph{Proceedings of the 2018 Conference on Empirical Methods in
  Natural Language Processing}, pages 2322--2332, Brussels, Belgium.
  Association for Computational Linguistics.

\bibitem[{Prakken et~al.(2020)}]{prakken2020persuasive}
H~Prakken et~al. 2020.
\newblock A persuasive chatbot using a crowd-sourced argument graph and
  concerns.
\newblock \emph{Computational Models of Argument: Proceedings of COMMA 2020},
  326:9.

\bibitem[{See et~al.(2017)See, Liu, and Manning}]{see-etal-2017-get}
Abigail See, Peter~J. Liu, and Christopher~D. Manning. 2017.
\newblock \href {https://doi.org/10.18653/v1/P17-1099} {Get to the point:
  Summarization with pointer-generator networks}.
\newblock In \emph{Proceedings of the 55th Annual Meeting of the Association
  for Computational Linguistics (Volume 1: Long Papers)}, pages 1073--1083,
  Vancouver, Canada. Association for Computational Linguistics.

\bibitem[{Shi et~al.(2020)Shi, Wang, Oh, Zhang, Sahay, and Yu}]{Shi2020}
Weiyan Shi, Xuewei Wang, Yoo~Jung Oh, Jingwen Zhang, Saurav Sahay, and Zhou Yu.
  2020.
\newblock \href {https://doi.org/10.1145/3313831.3376843} {Effects of
  persuasive dialogues: Testing bot identities and inquiry strategies}.
\newblock CHI '20, page 1–13. Association for Computing Machinery.

\bibitem[{Shuster et~al.(2021{\natexlab{a}})Shuster, Poff, Chen, Kiela, and
  Weston}]{shuster2021retrieval}
Kurt Shuster, Spencer Poff, Moya Chen, Douwe Kiela, and Jason Weston.
  2021{\natexlab{a}}.
\newblock Retrieval augmentation reduces hallucination in conversation.
\newblock \emph{arXiv preprint arXiv:2104.07567}.

\bibitem[{Shuster et~al.(2021{\natexlab{b}})Shuster, Poff, Chen, Kiela, and
  Weston}]{shuster-etal-2021-retrieval-augmentation}
Kurt Shuster, Spencer Poff, Moya Chen, Douwe Kiela, and Jason Weston.
  2021{\natexlab{b}}.
\newblock \href {https://aclanthology.org/2021.findings-emnlp.320} {Retrieval
  augmentation reduces hallucination in conversation}.
\newblock In \emph{Findings of the Association for Computational Linguistics:
  EMNLP 2021}, pages 3784--3803, Punta Cana, Dominican Republic. Association
  for Computational Linguistics.

\bibitem[{Slonim et~al.(2021)Slonim, Bilu, Alzate, Bar-Haim, Bogin, Bonin,
  Choshen, Cohen-Karlik, Dankin, Edelstein et~al.}]{slonim2021autonomous}
Noam Slonim, Yonatan Bilu, Carlos Alzate, Roy Bar-Haim, Ben Bogin, Francesca
  Bonin, Leshem Choshen, Edo Cohen-Karlik, Lena Dankin, Lilach Edelstein,
  et~al. 2021.
\newblock An autonomous debating system.
\newblock \emph{Nature}, 591(7850):379--384.

\bibitem[{Stanley et~al.(2020)Stanley, Whitehead, Sinnott-Armstrong, and
  Seli}]{STANLEY2020104030}
Matthew~L. Stanley, Peter~S. Whitehead, Walter Sinnott-Armstrong, and Paul
  Seli. 2020.
\newblock \href
  {https://www.sciencedirect.com/science/article/pii/S002210312030370X}
  {Exposure to opposing reasons reduces negative impressions of ideological
  opponents}.
\newblock \emph{Journal of Experimental Social Psychology}, 91:104030.

\bibitem[{Tan et~al.(2016)Tan, Niculae, Danescu-Niculescu-Mizil, and
  Lee}]{tan2016winning}
Chenhao Tan, Vlad Niculae, Cristian Danescu-Niculescu-Mizil, and Lillian Lee.
  2016.
\newblock Winning arguments: Interaction dynamics and persuasion strategies in
  good-faith online discussions.
\newblock In \emph{Proceedings of the 25th international conference on world
  wide web}, pages 613--624.

\bibitem[{Vecchi et~al.(2021)Vecchi, Falk, Jundi, and
  Lapesa}]{vecchi-etal-2021-towards}
Eva~Maria Vecchi, Neele Falk, Iman Jundi, and Gabriella Lapesa. 2021.
\newblock \href {https://doi.org/10.18653/v1/2021.acl-long.107} {Towards
  argument mining for social good: A survey}.
\newblock In \emph{Proceedings of the 59th Annual Meeting of the Association
  for Computational Linguistics and the 11th International Joint Conference on
  Natural Language Processing (Volume 1: Long Papers)}, pages 1338--1352,
  Online. Association for Computational Linguistics.

\bibitem[{Wang et~al.(2019)Wang, Shi, Kim, Oh, Yang, Zhang, and
  Yu}]{wang-etal-2019-persuasion}
Xuewei Wang, Weiyan Shi, Richard Kim, Yoojung Oh, Sijia Yang, Jingwen Zhang,
  and Zhou Yu. 2019.
\newblock \href {https://doi.org/10.18653/v1/P19-1566} {Persuasion for good:
  Towards a personalized persuasive dialogue system for social good}.
\newblock In \emph{Proceedings of the 57th Annual Meeting of the Association
  for Computational Linguistics}, pages 5635--5649, Florence, Italy.
  Association for Computational Linguistics.

\bibitem[{Yeomans et~al.(2020)Yeomans, Minson, Collins, Chen, and
  Gino}]{YEOMANS2020131}
Michael Yeomans, Julia Minson, Hanne Collins, Frances Chen, and Francesca Gino.
  2020.
\newblock \href
  {https://www.sciencedirect.com/science/article/pii/S0749597819303425}
  {Conversational receptiveness: Improving engagement with opposing views}.
\newblock \emph{Organizational Behavior and Human Decision Processes},
  160:131--148.

\bibitem[{Zhang et~al.(2016)Zhang, Kumar, Ravi, and
  Danescu-Niculescu-Mizil}]{zhang-etal-2016-conversational}
Justine Zhang, Ravi Kumar, Sujith Ravi, and Cristian Danescu-Niculescu-Mizil.
  2016.
\newblock \href {https://doi.org/10.18653/v1/N16-1017} {Conversational flow in
  {O}xford-style debates}.
\newblock In \emph{Proceedings of the 2016 Conference of the North {A}merican
  Chapter of the Association for Computational Linguistics: Human Language
  Technologies}, pages 136--141, San Diego, California. Association for
  Computational Linguistics.

\bibitem[{Zhao et~al.(2020{\natexlab{a}})Zhao, Wu, Tao, Xu, Zhao, and
  Yan}]{Zhao2020Low-Resource}
Xueliang Zhao, Wei Wu, Chongyang Tao, Can Xu, Dongyan Zhao, and Rui Yan.
  2020{\natexlab{a}}.
\newblock \href {https://openreview.net/forum?id=rJeIcTNtvS} {Low-resource
  knowledge-grounded dialogue generation}.
\newblock In \emph{International Conference on Learning Representations}.

\bibitem[{Zhao et~al.(2020{\natexlab{b}})Zhao, Wu, Xu, Tao, Zhao, and
  Yan}]{zhao-etal-2020-knowledge-grounded}
Xueliang Zhao, Wei Wu, Can Xu, Chongyang Tao, Dongyan Zhao, and Rui Yan.
  2020{\natexlab{b}}.
\newblock \href {https://doi.org/10.18653/v1/2020.emnlp-main.272}
  {Knowledge-grounded dialogue generation with pre-trained language models}.
\newblock In \emph{Proceedings of the 2020 Conference on Empirical Methods in
  Natural Language Processing (EMNLP)}, pages 3377--3390, Online. Association
  for Computational Linguistics.

\bibitem[{Zhou et~al.(2018)Zhou, Prabhumoye, and
  Black}]{zhou-etal-2018-dataset}
Kangyan Zhou, Shrimai Prabhumoye, and Alan~W Black. 2018.
\newblock \href {https://doi.org/10.18653/v1/D18-1076} {A dataset for document
  grounded conversations}.
\newblock In \emph{Proceedings of the 2018 Conference on Empirical Methods in
  Natural Language Processing}, pages 708--713, Brussels, Belgium. Association
  for Computational Linguistics.

\end{thebibliography}
\bibliographystyle{acl_natbib}

\appendix
\clearpage
\section{Instructions for Participants}
\label{sec:appendix-hp}
This study should take you no more than 25 minutes to complete, and you will be reimbursed £$3.75$ by Prolific. 

\noindent We are developing automated dialogue agents (“chatbots”) to have constructive conversations with people on a variety of topics. 

\noindent To train the automated system we need participants to “chat” with the dialogue agents.

\noindent Today you will be asked to join a chatroom to discuss the topic of veganism. You can interact with the system and discuss the topic for between $15$ and $20$ minutes. You can type anything you wish, as long as it is on the topic of veganism. 

\noindent The chat is only between you and the dialogue system, no other people will be joining your chatroom. 

\noindent The chatbot you interact with uses statements provided by other people online. These statements HAVE NOT BEEN FACT-CHECKED. 

\noindent \textbf{Your chat will be entirely anonymous and you WILL NOT have to give any personal information at any time.} 

\clearpage
\section{Feature Correlations with OUM Scores}
\label{sec:features-app}
\begin{table}[H]
    \begin{tabular}{ll|c}
\multicolumn{2}{c|}{Feature} & Spearman's $\rho$ \\ \hline
1. & length & 0.0 \\
2. & wizard turns & 0.018 \\
3. & participant turns & -0.017 \\
4. & wizard questions & -0.071 \\
5. & args from argument-base & 0.083 \\
6. & edited args & -0.123 \\
7. & pro/con ratio & 0.103 \\
8. & please & 0.0 \\
9. & please start & -0.059 \\
10.& has hedge & -0.026 \\
11.& indirect (btw) & 0.0 \\
12. & hedges & 0.028 \\
13. & factuality & 0.0 \\
14. & deference & -0.149 \\
15. & gratitude & -0.027 \\
16. & apologizing & 0.076 \\
17. & 1st person plural & 0.149 \\
18. & 1st person & -0.094 \\
19. & 1st person start & -0.075 \\
20. & 2nd person & -0.102 \\
21. & 2nd person start & 0.066 \\
22. & indirect greeting & -0.083 \\
23. & direct question & -0.096 \\
24. & direct start &. 0.045 \\
25. & positive lexicon & 0.181 \\
26. & negative lexicon & -0.02 \\
29. & subjunctive & -0.181 \\
30. & indicative & -0.024 \\ \hline
    \end{tabular}
    \caption{Spearman's correlation between dialogue features and OUM scores for the good reasons question. The features from 8 to 30 are politeness features from~\citet{danescu-niculescu-mizil-etal-2013-computational} extracted by Convokit~\cite{chang-etal-2020-convokit}.}
    \label{tab:features}
\end{table}

\clearpage
\section{Ratings for Chat Experiences}
\label{sec:bots-experience}
\begin{figure}[H]
  \centering
  \includegraphics[width=1\linewidth]{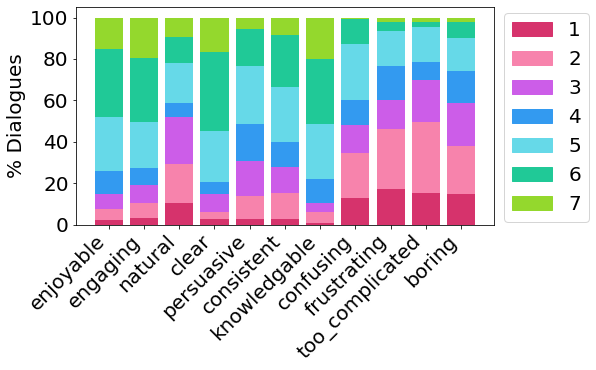}
  \caption{Ratings for chat experiences for the \textbf{argu-bot}. The y-axis corresponds to the proportion of the dialogues, the x-axis corresponds to chat experiences and the different colors refer to the ratings on the $7$-point Likert scale, where $1$=strongly disagree and $7$=strongly agree.}
  \label{fig:sub1}
\end{figure}%
\begin{figure}[H]
  \centering
  \includegraphics[width=1\linewidth]{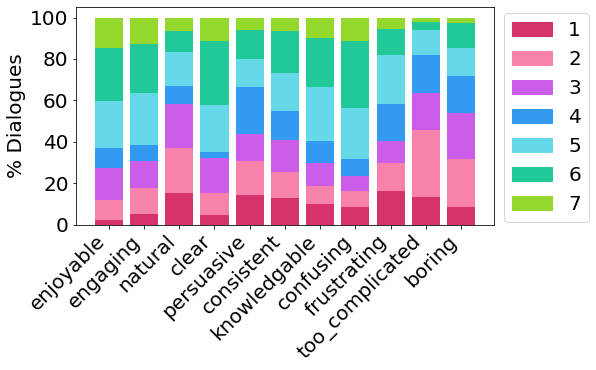}
  \caption{Ratings for chat experiences for the \textbf{wiki-bot}. The y-axis corresponds to the proportion of the dialogues, the x-axis corresponds to chat experiences and the different colors refer to the ratings on the $7$-point Likert scale, where $1$=strongly disagree and $7$=strongly agree.}
\label{fig:test}
  \label{fig:sub2}
\end{figure}

\end{document}